\documentclass[11pt]{article}

\usepackage[utf8]{inputenc}
\usepackage[T1]{fontenc}
\usepackage{lmodern}
\usepackage[margin=1in]{geometry}
\usepackage{booktabs}
\usepackage{graphicx}
\usepackage{natbib}
\usepackage{hyperref}
\usepackage{xcolor}
\usepackage{amsmath}
\usepackage{multirow}
\usepackage{tabularx}
\usepackage{caption}
\usepackage{subcaption}
\usepackage{enumitem}
\usepackage{array}

\hypersetup{
  colorlinks=true,
  linkcolor=blue!60!black,
  citecolor=blue!60!black,
  urlcolor=blue!60!black
}

\newcommand{\bench}{\textsc{BrainBench}}
\newcommand{\pp}{\,pp}

\title{\textbf{BrainBench: Exposing the Commonsense Reasoning Gap\\in Large Language Models}}

\author{
  Yuzhe Tang\\
  Georgia Institute of Technology\\
  \texttt{ytang352@gatech.edu}
}

\date{}

\begin{document}

\maketitle

\begin{abstract}
Large language models (LLMs) achieve impressive scores on standard benchmarks yet routinely fail questions that any human would answer correctly in seconds.
We introduce \bench{}, a benchmark of 100 brainteaser questions spanning 20 carefully designed categories, each targeting a specific commonsense reasoning failure mode in LLMs.
Categories range from \emph{implicit physical constraints} (``Should I walk or drive my rental car to the return lot?'') to \emph{semantic scope tricks} and \emph{default assumption hijacks}.
We evaluate eight frontier models---four from the Claude family and four from the GPT family---using a zero-shot protocol with 10 independent runs per question.
The best model, Claude Opus 4.6 with extended thinking, achieves only 80.3\% accuracy; the worst, GPT-4o, scores 39.7\%.
Even top-performing models exhibit a 6--16 percentage-point gap between accuracy and consistency, revealing stochastic reasoning.
Cross-lingual evaluation in Chinese shows most models degrade by 2--8\pp{}, confirming that these failures reflect reasoning deficits rather than language-specific artifacts.
\bench{} provides a fine-grained diagnostic tool for identifying where and why LLMs substitute surface heuristics for genuine commonsense reasoning.
\end{abstract}

\section{Introduction}
\label{sec:intro}

Consider the following question: \emph{``I need to return my rental car. The rental agency is just across the street. Should I walk over or drive?''} Every human immediately recognizes that you must \emph{drive}---the car itself needs to be returned. Yet when this class of question was posed to leading AI models in the Opper.ai Car Wash Test \citep{opper2025carwash}, the majority recommended walking, treating the problem as a simple distance-optimization task while ignoring the implicit physical constraint that the object must travel with you.

This is not an isolated failure. A growing body of evidence suggests that large language models (LLMs), despite their remarkable performance on standard NLP benchmarks \citep{wang2019superglue, srivastava2023beyond}, systematically fail on questions requiring basic commonsense reasoning \citep{davis2015commonsense, marcus2020next}. These are not trick questions in the adversarial sense---they are questions that any adult human answers effortlessly, often without conscious deliberation. The difficulty lies not in the complexity of the reasoning but in the need to override a surface-level heuristic with a deeper understanding of how the physical or social world works.

Existing benchmarks capture fragments of this problem. The BRAINTEASER shared task at SemEval-2024 \citep{jiang2024brainteaser} focused on lateral thinking puzzles requiring creative reinterpretation. SimpleBench \citep{simplebench2024} tested spatial, temporal, and social reasoning. HellaSwag \citep{zellers2019hellaswag} and PIQA \citep{bisk2020piqa} evaluate commonsense through sentence completion and physical intuition. However, none of these benchmarks provides a systematic \emph{taxonomy of failure modes}---a categorization of the specific types of reasoning traps that LLMs fall into, organized by the underlying cognitive mechanism.

We introduce \bench{}, a benchmark designed to fill this gap. Our contributions are:

\begin{enumerate}[leftmargin=*, itemsep=2pt]
  \item A \textbf{taxonomy of 20 commonsense reasoning failure categories}, each defined by a specific cognitive trap (e.g., broken-device self-reference, wrong vantage point, default assumption hijack), the surface heuristic that LLMs follow, and why that heuristic fails.
  \item A \textbf{benchmark of 100 questions} (5 per category), designed so that each question is trivially easy for humans but exploits a known LLM reasoning gap.
  \item A \textbf{comprehensive evaluation of 8 frontier models} across two model families (Claude and GPT), including extended-thinking variants, using a zero-shot protocol with 10 independent runs per question to measure both accuracy and consistency.
  \item A \textbf{cross-lingual evaluation} in Chinese, revealing that reasoning failures persist across languages and are not artifacts of English-specific phrasing.
\end{enumerate}

Our key findings reveal a clear three-tier structure in commonsense reasoning ability: Claude-family models cluster at 74--80\% accuracy, GPT-5.4 models at 70--74\%, and GPT-4o models at approximately 40\%. The hardest categories---\emph{implicit physical constraint} and \emph{wrong vantage point}---average only 40\% accuracy across all models, meaning that even frontier LLMs perform near chance on questions any child could answer. Extended thinking provides a modest benefit of approximately 3\pp{} overall but actually \emph{hurts} performance on certain category types, suggesting that more computation does not uniformly translate to better commonsense reasoning.

\section{Related Work}
\label{sec:related}

\paragraph{Commonsense reasoning benchmarks.}
Evaluating commonsense in AI systems has a long history \citep{davis2015commonsense}. Modern benchmarks approach the problem from different angles. CommonsenseQA \citep{talmor2019commonsenseqa} tests knowledge-intensive commonsense via multiple-choice questions grounded in ConceptNet. PIQA \citep{bisk2020piqa} focuses on physical intuition (``Which is a better way to cut a piece of glass?''). HellaSwag \citep{zellers2019hellaswag} uses adversarially filtered sentence completions to test situational commonsense. RiddleSense \citep{lin2021riddlesense} probes creative commonsense through riddles. The RAINBOW benchmark \citep{lourie2021unicorn} unifies several commonsense tasks into a multitask framework. While these benchmarks have driven progress, state-of-the-art LLMs now achieve near-human performance on most of them \citep{openai2023gpt4}, raising the question of whether they measure genuine reasoning or pattern-matching ability. \bench{} is designed to remain challenging precisely because it targets the gap between surface-level pattern matching and genuine understanding.

\paragraph{The Car Wash Test.}
\citet{opper2025carwash} introduced a viral evaluation in which 53 AI models were asked whether to walk or drive to a nearby car wash. The question exploits the same implicit physical constraint as our Category 1: the car itself must be present at the car wash. The majority of tested models recommended walking, revealing that they optimized for distance rather than reasoning about the task's physical requirements. The Car Wash Test demonstrated that a single well-designed question can expose fundamental reasoning gaps, but as a single-question evaluation it cannot characterize the breadth of failure modes. \bench{} extends this insight into a systematic taxonomy.

\paragraph{BRAINTEASER and lateral thinking.}
The SemEval-2024 BRAINTEASER task \citep{jiang2024brainteaser} evaluated LLMs on sentence puzzles and word puzzles that require defying default commonsense associations. Their work demonstrated that LLMs struggle with questions requiring lateral thinking---breaking away from the most obvious interpretation. \bench{} complements this by providing a finer-grained categorization: rather than grouping all lateral thinking failures together, we decompose them into 20 distinct mechanisms (e.g., semantic scope tricks vs.\ default assumption hijacks vs.\ answer hiding in plain sight).

\paragraph{SimpleBench.}
SimpleBench \citep{simplebench2024} evaluates LLMs on questions requiring spatial reasoning, temporal reasoning, and social intelligence. It shares our motivation of testing supposedly ``simple'' tasks, but focuses on a coarser categorization (spatial, temporal, social) rather than the specific cognitive traps that cause failures.

\paragraph{LLM reasoning and chain-of-thought.}
Chain-of-thought prompting \citep{wei2022chain} and extended thinking modes have been shown to improve performance on complex reasoning tasks. Recent work \citep{mahowald2024dissociating} argues that language competence and reasoning ability are dissociable in LLMs---models can exhibit fluent language production while lacking robust reasoning. Our thinking-mode analysis (Section~\ref{sec:results}) provides empirical evidence for this dissociation: extended thinking yields only modest improvements on commonsense brainteasers and actually degrades performance on certain categories.

\section{The BrainBench Dataset}
\label{sec:dataset}

\bench{} consists of 100 questions organized into 20 categories, with 5 questions per category. Each category targets a specific commonsense reasoning failure mode, defined by (a)~the \emph{core trap}: the surface-level heuristic that LLMs follow; (b)~the \emph{why}: the reason this heuristic leads to the wrong answer; and (c)~the \emph{correct reasoning}: the commonsense knowledge needed to override the heuristic. Table~\ref{tab:categories} presents the full taxonomy.

\subsection{Category Design}

The 20 categories were identified through a systematic analysis of LLM failure patterns on commonsense questions reported in prior work \citep{opper2025carwash, jiang2024brainteaser, simplebench2024}, online AI failure compilations, and our own exploratory testing. We organized these failures by the \emph{type of cognitive shortcut} that produces the error, yielding a taxonomy that spans physical reasoning (Categories 1--5), linguistic/semantic reasoning (Categories 6--7, 15, 16, 19--20), logical reasoning (Categories 8--9, 12--13, 18), and social/pragmatic reasoning (Categories 10--11, 14, 17). Table~\ref{tab:categories} provides the complete taxonomy.

\begin{table}[t]
\centering
\caption{The 20 \bench{} categories. Each category targets a specific reasoning trap. Avg.\ Acc.\ is the mean accuracy across all 8 models.}
\label{tab:categories}
\small
\begin{tabular}{@{}r l p{6.2cm} r@{}}
\toprule
\textbf{\#} & \textbf{Category} & \textbf{Core Trap} & \textbf{Avg.\ Acc.} \\
\midrule
1 & Implicit physical constraint & Surface cue (short distance) overrides need for object to be present & 40.0\% \\
2 & Broken/dead device self-ref. & Recommends broken device for the task it cannot perform & 60.8\% \\
3 & Wrong test conditions & Favors immediate action; ignores that conditions invalidate the test & 63.3\% \\
4 & Self-defeating action & Direct action sounds efficient but physically undermines the goal & 72.5\% \\
5 & Wrong vantage point & Uses nearest observation tool; ignores spatial geometry & 40.0\% \\
6 & Embedded false premise & Answers within a premise that should be rejected outright & 75.8\% \\
7 & Semantic scope trick & Interprets a key word with wrong scope (e.g., ``exactly'' vs.\ ``at least'') & 50.0\% \\
8 & State/identity tracking & Applies the event label rather than tracking resulting state & 80.0\% \\
9 & Temporal impossibility & Endorses action whose window has already closed & 78.3\% \\
10 & Default assumption hijack & Fills in unstated details with stereotypical defaults & 51.7\% \\
11 & Red herring overload & Over-processes irrelevant detail; misses trivially simple answer & 70.0\% \\
12 & Quantity/counting illusion & Performs the obvious arithmetic, which is wrong & 81.7\% \\
13 & Scale/growth intuition fail. & Linear intuition underestimates exponential growth & 95.0\% \\
14 & Naive physics error & Reproduces folk-physics misconceptions from training data & 73.3\% \\
15 & Pragmatic/social intent & Answers literally; misses that the utterance is a request or sarcasm & 56.7\% \\
16 & Framing/anchoring trap & Follows the pattern the question's framing establishes & 70.8\% \\
17 & Goal--means mismatch & Endorses action associated with goal that does not causally achieve it & 78.3\% \\
18 & Circular dependency & Recommends tool whose use requires the thing it would provide & 73.3\% \\
19 & Answer hiding in plain sight & Searches for complex solution; overlooks answer stated in the question & 59.2\% \\
20 & Negation/exception logic & Miscounts nested negations; gets polarity backwards & 60.8\% \\
\bottomrule
\end{tabular}
\end{table}

\subsection{Question Design}
\label{sec:question_design}

Each question in \bench{} is designed to satisfy three criteria: (1)~\textbf{trivially easy for humans}---an average adult should answer correctly within seconds; (2)~\textbf{trap-specific}---the question activates the specific surface heuristic defined by its category; and (3)~\textbf{unambiguous}---the correct answer is objectively determinable, not a matter of opinion.

Questions vary in surface complexity. Some are deliberately short (``My car battery is dead. Should I drive to the auto parts store or have someone bring me a battery?''), while others embed the trap within a longer, more naturalistic scenario featuring expert-sounding but flawed advice. The longer format tests whether models can identify the critical flaw amid plausible-sounding detail---a skill that requires genuine comprehension rather than pattern matching.

\subsection{Illustrative Examples}

We highlight three examples that illustrate different failure mechanisms:

\paragraph{Example 1: Implicit physical constraint.}
\emph{``I need to get my boat repaired. The repair dock is 100 meters along the shore. Should I walk or sail?''} The correct answer is \emph{sail}---the boat must be at the dock. Models that recommend walking optimize for the easier mode of transport without reasoning about the physical constraint that the repair target must be present.

\paragraph{Example 2: Broken/dead device self-reference.}
\emph{``My internet connection is down. Should I google the troubleshooting steps or call my ISP?''} The correct answer is \emph{call}---googling requires the very internet connection that is down. This exploits the model's association between ``troubleshooting'' and ``search online.''

\paragraph{Example 3: Default assumption hijack.}
Questions in this category exploit cultural stereotypes or default scripts (e.g., assuming a surgeon is male, or that a dark scene implies nighttime). The model fills in unstated details with statistically dominant patterns rather than considering alternatives consistent with the stated facts.

\section{Experimental Setup}
\label{sec:setup}

\subsection{Models}

We evaluate eight frontier LLMs spanning two major model families, as shown in Table~\ref{tab:models}. This selection covers a range of model sizes within each family and includes extended-thinking variants for both the Claude and GPT families.

\begin{table}[t]
\centering
\caption{Models evaluated. ``Think'' denotes extended thinking (reasoning) mode.}
\label{tab:models}
\small
\begin{tabular}{@{}l l l@{}}
\toprule
\textbf{Model} & \textbf{Provider} & \textbf{Variant} \\
\midrule
Claude Haiku 4.5 & Anthropic & Standard \\
Claude Sonnet 4.6 & Anthropic & Standard \\
Claude Opus 4.6 & Anthropic & Standard \\
Claude Opus 4.6 Think & Anthropic & Extended thinking \\
GPT-4o Mini & OpenAI & Standard \\
GPT-4o & OpenAI & Standard \\
GPT-5.4 & OpenAI & Standard \\
GPT-5.4 Think & OpenAI & Extended thinking \\
\bottomrule
\end{tabular}
\end{table}

\subsection{Evaluation Protocol}

We adopt a strict zero-shot protocol: each question is sent to the model as a standalone prompt with no system prompt, no few-shot examples, and no chain-of-thought instructions. This isolates the model's default reasoning behavior, following the methodology of the Car Wash Test \citep{opper2025carwash}.

Each question is evaluated \textbf{10 independent times} per model (300 total responses per model across 100 questions). Multiple runs allow us to distinguish between reliable reasoning and stochastic correctness---a model that answers correctly 3 out of 10 times is meaningfully different from one that answers correctly 10 out of 10 times.

\subsection{Automated Judging}

Responses are evaluated by an LLM judge that compares each response against the ground-truth answer. The judge makes a binary determination (correct/incorrect) based on whether the response's conclusion matches the ground-truth reasoning, regardless of phrasing or verbosity. We use a strong model as the judge and validated its agreement with human annotators.

\subsection{Metrics}

We report two primary metrics:

\begin{itemize}[leftmargin=*, itemsep=2pt]
  \item \textbf{Accuracy}: the fraction of runs in which the model produces the correct answer, aggregated across all 10 runs per question.
  \item \textbf{Consistency} (Reliability): the fraction of questions for which the model produces the correct answer in \emph{all} 10 runs. A model with 80\% accuracy but 50\% consistency gets many questions right sometimes but wrong other times, indicating stochastic reasoning rather than robust understanding.
\end{itemize}

The gap between accuracy and consistency captures \emph{reasoning reliability}: how much a model's correctness depends on the particular sample from its output distribution.

\subsection{Cross-Lingual Evaluation}

To test whether \bench{} failures are English-specific or reflect deeper reasoning deficits, we translate the full benchmark into Chinese and evaluate all eight models under the same protocol. Translation preserves the semantic content and trap structure of each question while adapting cultural references where necessary.

\section{Results}
\label{sec:results}

\subsection{Overall Performance}

\begin{figure}[t]
  \centering
  \includegraphics[width=\columnwidth]{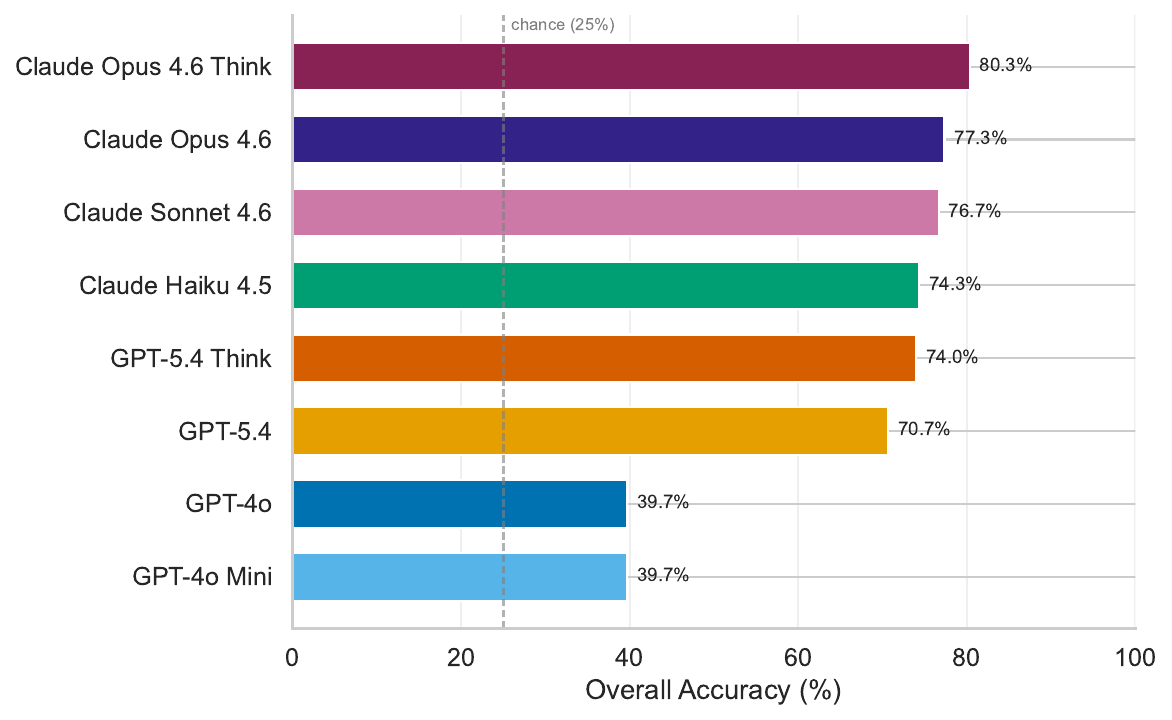}
  \caption{Overall accuracy (\%) on \bench{} (English). Models are ranked by accuracy. Error bars indicate the range between accuracy and consistency, capturing reasoning reliability.}
  \label{fig:overall}
\end{figure}

Figure~\ref{fig:overall} presents the overall accuracy ranking. A clear three-tier structure emerges:

\begin{itemize}[leftmargin=*, itemsep=2pt]
  \item \textbf{Tier 1} (74--80\%): All four Claude models, led by Claude Opus 4.6 Think at 80.3\%.
  \item \textbf{Tier 2} (70--74\%): GPT-5.4 and GPT-5.4 Think, at 70.7\% and 74.0\% respectively.
  \item \textbf{Tier 3} ($\sim$40\%): GPT-4o and GPT-4o Mini, both at 39.7\%.
\end{itemize}

The 40.7\pp{} spread between the best and worst models demonstrates that \bench{} discriminates effectively between model capabilities. Notably, even the smallest Claude model (Haiku 4.5, 74.3\%) outperforms the largest non-thinking GPT model (GPT-5.4, 70.7\%) by 3.6\pp{}, suggesting fundamental differences in commonsense reasoning between model families.

Table~\ref{tab:leaderboard} provides the complete leaderboard with both accuracy and consistency scores.

\begin{table}[t]
\centering
\caption{Overall leaderboard on \bench{} (English). Consistency measures the fraction of questions answered correctly in all 10 runs. Gap = Accuracy $-$ Consistency.}
\label{tab:leaderboard}
\small
\begin{tabular}{@{}r l c c c c@{}}
\toprule
\textbf{Rank} & \textbf{Model} & \textbf{Accuracy} & \textbf{Consistency} & \textbf{Gap} & \textbf{Correct/Total} \\
\midrule
1 & Claude Opus 4.6 Think & 80.3\% & 74.0\% & 6.3\pp{} & 241/300 \\
2 & Claude Opus 4.6 & 77.3\% & 71.0\% & 6.3\pp{} & 232/300 \\
3 & Claude Sonnet 4.6 & 76.7\% & 69.0\% & 7.7\pp{} & 230/300 \\
4 & Claude Haiku 4.5 & 74.3\% & 58.0\% & 16.3\pp{} & 223/300 \\
5 & GPT-5.4 Think & 74.0\% & 64.0\% & 10.0\pp{} & 222/300 \\
6 & GPT-5.4 & 70.7\% & 63.0\% & 7.7\pp{} & 212/300 \\
7 & GPT-4o Mini & 39.7\% & 24.0\% & 15.7\pp{} & 119/300 \\
8 & GPT-4o & 39.7\% & 27.0\% & 12.7\pp{} & 119/300 \\
\bottomrule
\end{tabular}
\end{table}

\subsection{Category-Level Analysis}

\begin{figure}[t]
  \centering
  \includegraphics[width=\columnwidth]{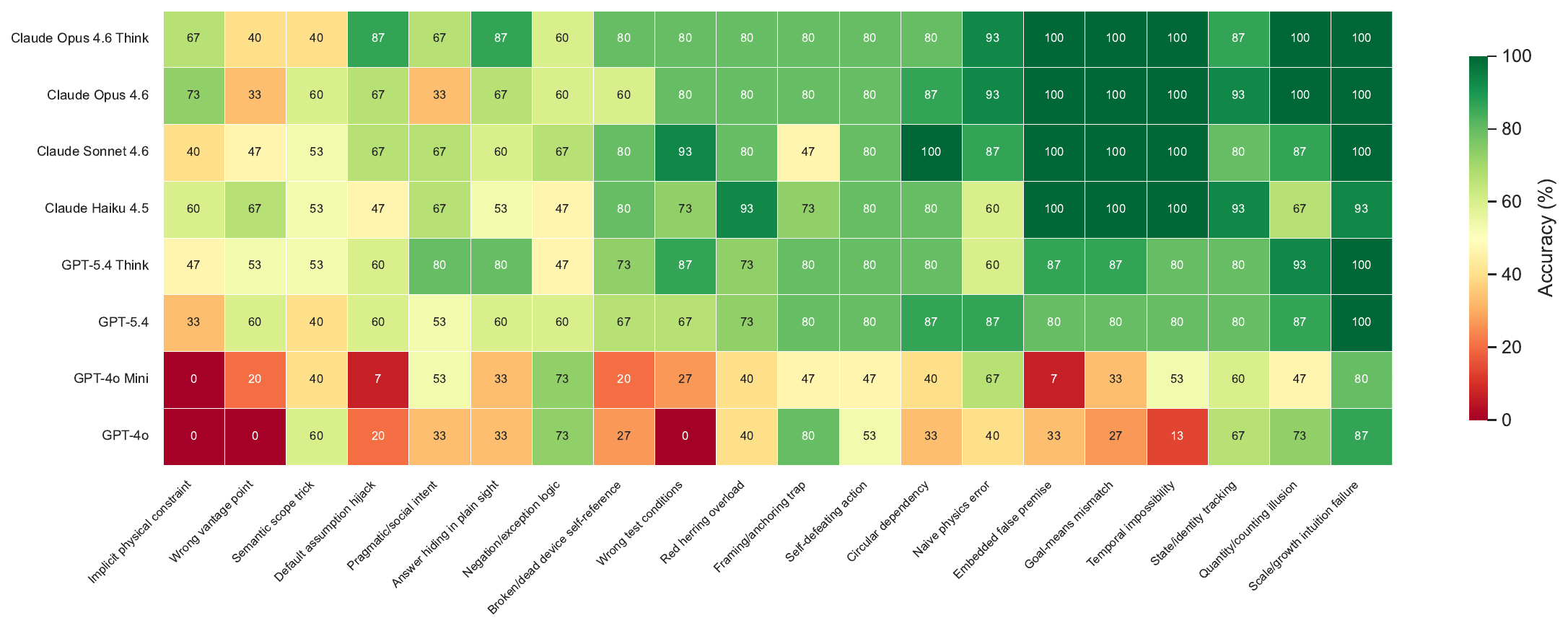}
  \caption{Accuracy heatmap across models and categories. Categories are sorted by average difficulty (hardest at left). Darker cells indicate lower accuracy.}
  \label{fig:heatmap}
\end{figure}

Figure~\ref{fig:heatmap} shows the category-level heatmap, revealing striking variation in difficulty across categories.

\paragraph{Hardest categories.}
Two categories stand out as exceptionally difficult, averaging only 40\% accuracy across all models:

\begin{enumerate}[leftmargin=*, itemsep=2pt]
  \item \textbf{Implicit physical constraint} (40.0\%): Questions where a surface cue (short distance, convenience) triggers a default response that ignores the requirement for an object to be physically present. GPT-4o and GPT-4o Mini score 0\% on this category.
  \item \textbf{Wrong vantage point} (40.0\%): Questions about observation or perception where the target is not visible from the assumed position. GPT-4o scores 0\%.
\end{enumerate}

Three additional categories average below 55\%: \emph{semantic scope trick} (50.0\%), \emph{default assumption hijack} (51.7\%), and \emph{pragmatic/social intent} (56.7\%).

\paragraph{Easiest categories.}
\emph{Scale/growth intuition failure} is the easiest category at 95.0\% average accuracy, with even GPT-4o Mini achieving 80\%. This suggests that frontier models have largely internalized exponential growth reasoning, perhaps through extensive exposure to classic examples (e.g., lily pad doubling problems) in training data. \emph{Quantity/counting illusion} (81.7\%) and \emph{state/identity tracking} (80.0\%) are also relatively easy.

\paragraph{Category-model interactions.}
The heatmap reveals that no model is uniformly strong or weak across all categories. Claude Opus 4.6 Think achieves 100\% on five categories (embedded false premise, goal--means mismatch, temporal impossibility, quantity/counting illusion, scale/growth intuition failure) but scores only 40\% on wrong vantage point and 40\% on semantic scope trick. Similarly, GPT-4o scores 87\% on scale/growth intuition failure and 80\% on framing/anchoring trap, but 0\% on implicit physical constraint, wrong vantage point, and wrong test conditions. These asymmetric strengths and weaknesses make \bench{} a useful diagnostic tool for identifying specific reasoning gaps in individual models.

\subsection{Consistency Analysis}

The gap between accuracy and consistency reveals the reliability of each model's reasoning. The average gap across all models is 10.3\pp{}. The most consistent model is Claude Opus 4.6 Think, with a reliability-to-accuracy ratio of 0.92 (74.0\% consistency vs.\ 80.3\% accuracy). The least consistent is GPT-4o Mini, with a ratio of 0.61 (24.0\% consistency vs.\ 39.7\% accuracy).

Claude Haiku 4.5 presents an interesting case: despite ranking 4th in accuracy (74.3\%), it has the second-largest consistency gap (16.3\pp{}), suggesting that many of its correct answers are stochastic rather than reflecting robust reasoning. By contrast, GPT-5.4 has a relatively small gap (7.7\pp{}) despite lower accuracy (70.7\%), indicating that when it gets a question right, it does so reliably.

\subsection{Effect of Extended Thinking}

\begin{figure}[t]
  \centering
  \includegraphics[width=\columnwidth]{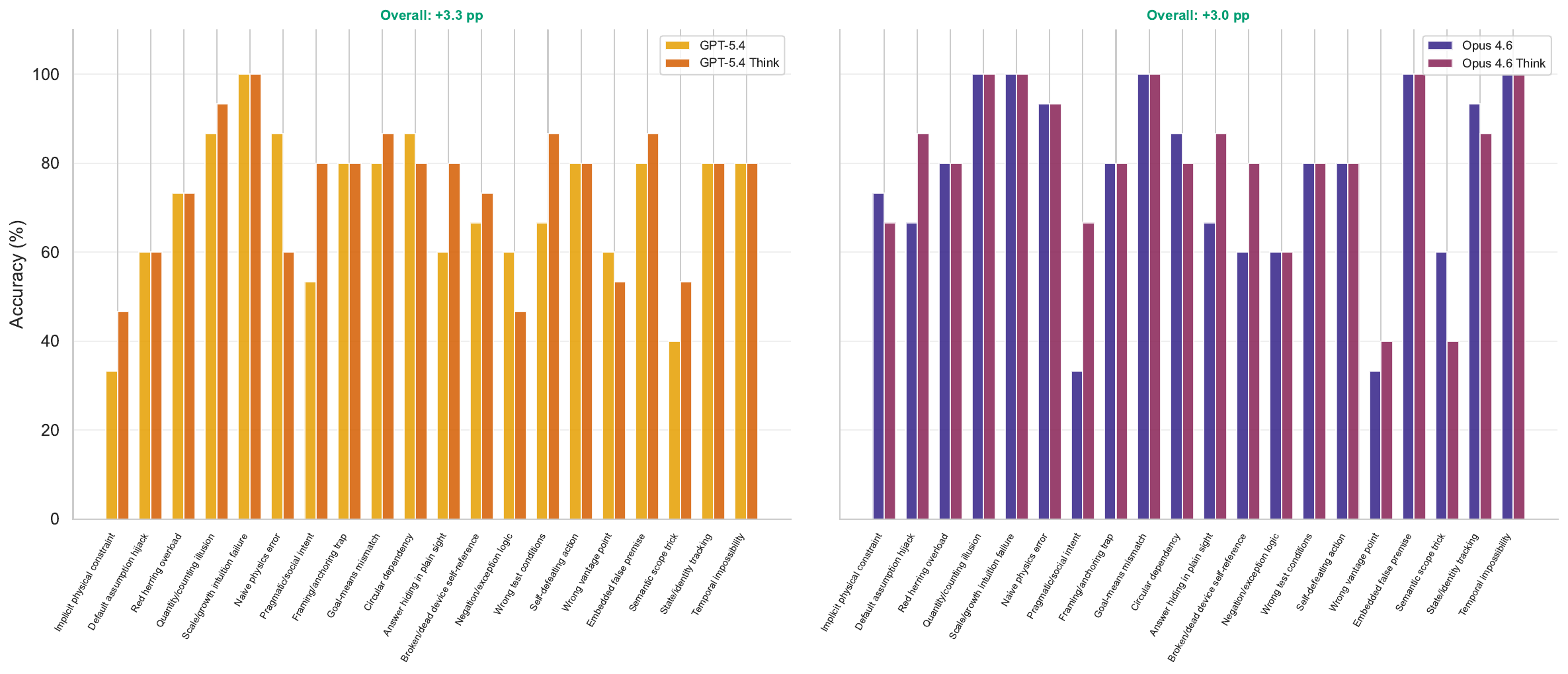}
  \caption{Accuracy comparison between standard and extended-thinking variants by category. Green bars indicate categories where thinking helps; red bars indicate where it hurts.}
  \label{fig:thinking}
\end{figure}

Figure~\ref{fig:thinking} compares standard and thinking-mode variants for both model families. Extended thinking provides a modest overall benefit: $+$3.3\pp{} for GPT-5.4 and $+$3.0\pp{} for Claude Opus 4.6. However, the effect is highly uneven across categories.

\paragraph{Where thinking helps.}
For GPT-5.4, the largest gains from thinking mode appear in categories requiring multi-step deliberation: \emph{pragmatic/social intent} ($+$26.7\pp{}), \emph{answer hiding in plain sight} ($+$20.0\pp{}), and \emph{wrong test conditions} ($+$20.0\pp{}). For Claude Opus 4.6, thinking helps most on \emph{pragmatic/social intent} ($+$33.3\pp{}) and \emph{default assumption hijack} ($+$20.0\pp{}).

\paragraph{Where thinking hurts.}
Strikingly, thinking mode \emph{degrades} performance on several categories. For GPT-5.4 Think, \emph{naive physics error} drops by 26.7\pp{} and \emph{negation/exception logic} by 13.3\pp{}. For Claude Opus 4.6 Think, \emph{semantic scope trick} drops by 20.0\pp{}. This pattern suggests that extended reasoning can lead to \emph{overthinking}---the model's initial intuition is correct, but deliberation introduces second-guessing or confabulated justifications for the wrong answer.

\subsection{Cross-Lingual Evaluation}

\begin{figure}[t]
  \centering
  \includegraphics[width=\columnwidth]{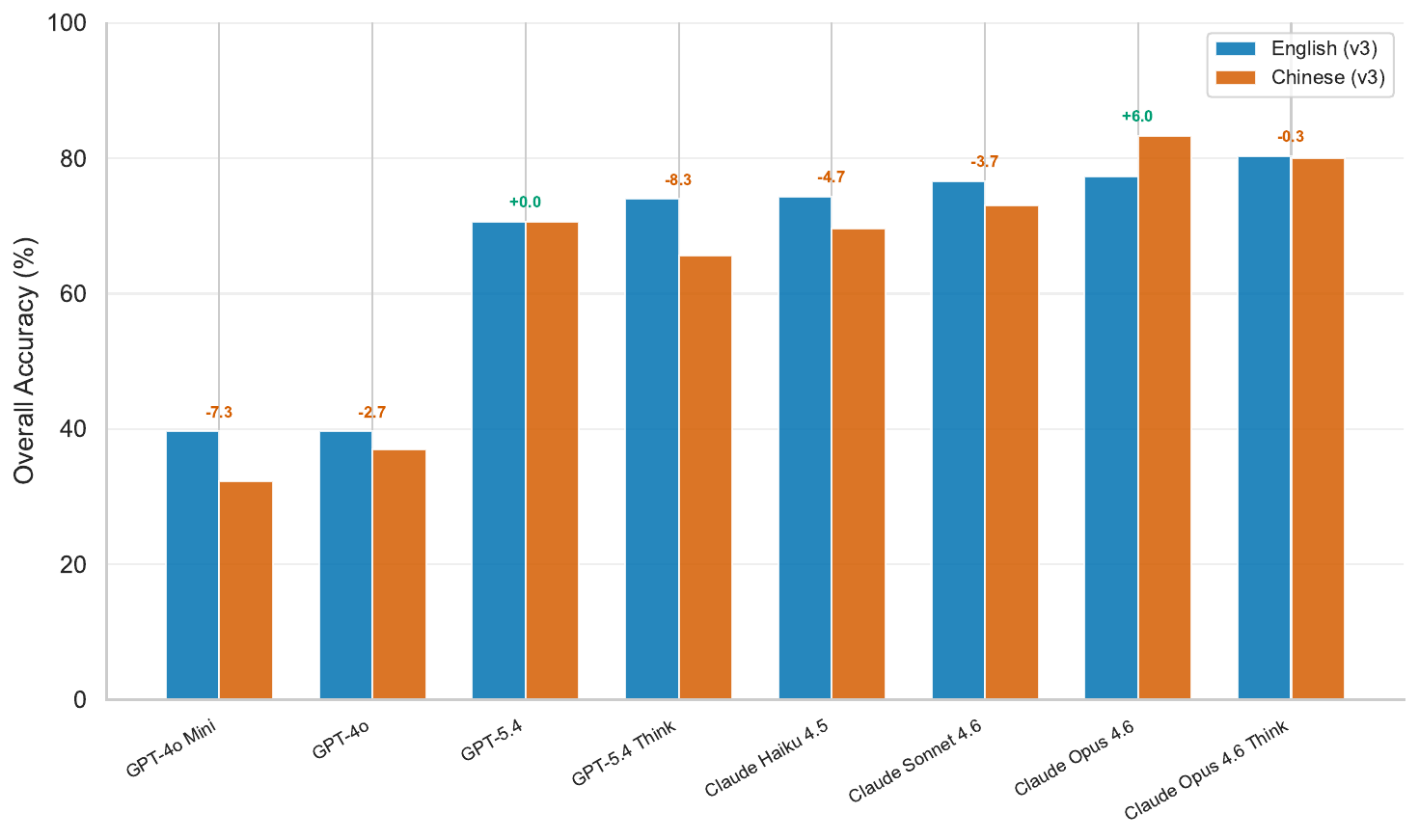}
  \caption{Accuracy comparison between English and Chinese versions of \bench{}.}
  \label{fig:crosslingual}
\end{figure}

Figure~\ref{fig:crosslingual} presents the English vs.\ Chinese comparison. The average accuracy across all models drops by 2.6\pp{} in Chinese (66.6\% English vs.\ 64.0\% Chinese), confirming that \bench{} failures are not artifacts of English phrasing.

Table~\ref{tab:crosslingual} shows the per-model breakdown. Most models degrade in Chinese, with GPT-5.4 Think showing the largest drop ($-$8.3\pp{}) and GPT-4o Mini declining by $-$7.3\pp{}. A notable exception is Claude Opus 4.6, which actually \emph{improves} by $+$6.0\pp{} in Chinese (83.3\% vs.\ 77.3\% in English), achieving the highest score of any model in any language. This unexpected result may reflect differences in how this model processes Chinese-language framing or stronger cross-lingual reasoning transfer in its architecture.

\begin{table}[t]
\centering
\caption{Cross-lingual comparison: English vs.\ Chinese accuracy. $\Delta$ is Chinese minus English.}
\label{tab:crosslingual}
\small
\begin{tabular}{@{}l c c r l@{}}
\toprule
\textbf{Model} & \textbf{English} & \textbf{Chinese} & \textbf{$\Delta$} & \textbf{Direction} \\
\midrule
Claude Opus 4.6 Think & 80.3\% & 80.0\% & $-$0.3\pp{} & EN $>$ CN \\
Claude Opus 4.6 & 77.3\% & 83.3\% & $+$6.0\pp{} & CN $>$ EN \\
Claude Sonnet 4.6 & 76.7\% & 73.0\% & $-$3.7\pp{} & EN $>$ CN \\
Claude Haiku 4.5 & 74.3\% & 69.7\% & $-$4.7\pp{} & EN $>$ CN \\
GPT-5.4 Think & 74.0\% & 65.7\% & $-$8.3\pp{} & EN $>$ CN \\
GPT-5.4 & 70.7\% & 70.7\% & $\pm$0.0\pp{} & Equal \\
GPT-4o Mini & 39.7\% & 32.3\% & $-$7.3\pp{} & EN $>$ CN \\
GPT-4o & 39.7\% & 37.0\% & $-$2.7\pp{} & EN $>$ CN \\
\midrule
\textbf{Average} & \textbf{66.6\%} & \textbf{64.0\%} & $-$\textbf{2.6\pp{}} & \\
\bottomrule
\end{tabular}
\end{table}

\subsection{Model Family Comparison}

Within the GPT family, the most striking finding is the 31.0\pp{} jump from GPT-4o (39.7\%) to GPT-5.4 (70.7\%), suggesting substantial reasoning improvements in the newer generation. By contrast, GPT-4o and GPT-4o Mini perform identically (39.7\%), and extended thinking adds only 3.3\pp{} to GPT-5.4.

The Claude family shows a flatter progression: all four models fall within a 6.0\pp{} range (74.3\% to 80.3\%). Even the smallest Claude model (Haiku 4.5) outperforms GPT-5.4 by 3.6\pp{}. In head-to-head comparisons at matched tiers, Claude models outperform their GPT counterparts at every level, with advantages ranging from 6.3\pp{} (Opus Think vs.\ GPT-5.4 Think) to 37.0\pp{} (Sonnet vs.\ GPT-4o).

\section{Discussion}
\label{sec:discussion}

\subsection{Why Do Models Fail?}

The category-level results suggest a unifying explanation: \textbf{LLMs default to surface heuristics when the correct answer requires overriding a statistically dominant pattern}. The hardest categories are precisely those where the most ``obvious'' response---the one that would be correct in the majority of similar-sounding scenarios---is wrong in the specific case presented.

Consider implicit physical constraint (40\% average accuracy). In most real-world contexts, when a destination is nearby, walking \emph{is} the sensible choice. The model has learned this pattern from millions of examples. The brainteaser subverts this pattern by adding a constraint (the object must travel with you) that requires overriding the default. The model's failure is not a lack of knowledge---it ``knows'' that cars need to be present for car washes and repairs---but a failure to activate that knowledge in context.

This aligns with the dissociation between language competence and reasoning proposed by \citet{mahowald2024dissociating}: models can produce fluent, knowledgeable text while failing to apply that knowledge to novel situations that conflict with learned patterns.

\subsection{The Overthinking Paradox}

Our finding that extended thinking helps on some categories but hurts on others reveals what we term the \emph{overthinking paradox}. On categories like \emph{pragmatic/social intent}, where the correct answer requires stepping back from literal interpretation to consider social context, additional reasoning time helps the model reconsider its initial literal parse. But on categories like \emph{semantic scope trick} and \emph{naive physics error}, the model's first intuition is sometimes correct, and extended deliberation introduces opportunities to talk itself into the wrong answer.

This pattern has practical implications: deploying thinking modes for commonsense tasks may not yield uniform improvement, and the optimal strategy depends on the type of reasoning required.

\subsection{Cross-Lingual Robustness}

The modest average degradation in Chinese ($-$2.6\pp{}) confirms that \bench{} measures reasoning deficits that transcend language-specific surface features. The remarkable performance of Claude Opus 4.6 in Chinese ($+$6.0\pp{} over English) warrants further investigation. One hypothesis is that Chinese phrasing, being structurally different from English, disrupts some of the English-centric surface heuristics that models have over-learned, inadvertently improving performance by forcing the model into a less ``autopilot'' mode.

\subsection{Universally Hard and Easy Questions}

Three questions achieve 0\% mean accuracy across all 8 models, meaning \emph{no model answered correctly in any run}. These represent reasoning blind spots shared across architectures. Conversely, 10 questions achieve 100\% accuracy across all models. The coexistence of universally-solved and universally-failed questions within the same category (e.g., \emph{self-defeating action} contains both) suggests that difficulty depends not just on the category but on the specific way the trap is instantiated.

\subsection{Limitations}

Several limitations should be noted. First, our benchmark contains 100 questions---sufficient for category-level analysis but limited for fine-grained statistical tests within categories (5 questions each). Expanding the dataset would strengthen per-category conclusions. Second, we lack a formal human baseline; while these questions are designed to be trivially easy for humans, a controlled study with human participants would quantify the human--AI gap precisely. Third, our evaluation covers two model families (Claude and GPT); including additional families (Gemini, Llama, etc.) would broaden the findings. Fourth, the LLM judge, while validated, may introduce systematic biases in edge cases. Finally, our cross-lingual evaluation covers only Chinese; extending to additional languages would further test the language-independence of the findings.

\section{Conclusion}
\label{sec:conclusion}

We introduced \bench{}, a benchmark of 100 commonsense brainteaser questions organized into 20 categories, each targeting a specific reasoning failure mode in large language models. Our evaluation of eight frontier models reveals that even the best model achieves only 80.3\% accuracy on questions that are trivially easy for humans, with the hardest categories averaging 40\% accuracy---near chance.

The \bench{} taxonomy provides a diagnostic tool for identifying \emph{where} and \emph{why} LLMs fail at commonsense reasoning: not from a lack of world knowledge, but from an over-reliance on surface heuristics that are correct in most contexts but fail in precisely the situations where commonsense matters most. As LLMs are increasingly deployed in real-world settings that require genuine understanding of physical constraints, social norms, and logical dependencies, benchmarks like \bench{} serve as a necessary complement to standard evaluations---measuring not what models know, but whether they can apply what they know when it counts.

The dataset and evaluation code are publicly available.\footnote{URL redacted for review.}

\bibliographystyle{plainnat}
\bibliography{references}

\appendix
\section{Category Difficulty Rankings}
\label{app:difficulty}

Table~\ref{tab:difficulty} presents the full category difficulty ranking, sorted by average accuracy across all eight models.

\begin{table}[ht]
\centering
\caption{Categories ranked by average accuracy (hardest first).}
\label{tab:difficulty}
\small
\begin{tabular}{@{}r l c l l@{}}
\toprule
\textbf{Rank} & \textbf{Category} & \textbf{Avg Acc.} & \textbf{Hardest For} & \textbf{Easiest For} \\
\midrule
1 & Implicit physical constraint & 40.0\% & GPT-4o Mini (0\%) & Claude Opus 4.6 (73\%) \\
2 & Wrong vantage point & 40.0\% & GPT-4o (0\%) & Claude Haiku 4.5 (67\%) \\
3 & Semantic scope trick & 50.0\% & GPT-4o Mini (40\%) & GPT-4o (60\%) \\
4 & Default assumption hijack & 51.7\% & GPT-4o Mini (7\%) & Claude Opus Think (87\%) \\
5 & Pragmatic/social intent & 56.7\% & GPT-4o (33\%) & GPT-5.4 Think (80\%) \\
6 & Answer hiding in plain sight & 59.2\% & GPT-4o Mini (33\%) & Claude Opus Think (87\%) \\
7 & Negation/exception logic & 60.8\% & GPT-5.4 Think (47\%) & GPT-4o Mini (73\%) \\
8 & Broken/dead device self-ref. & 60.8\% & GPT-4o Mini (20\%) & Haiku/Sonnet (80\%) \\
9 & Wrong test conditions & 63.3\% & GPT-4o (0\%) & Claude Sonnet 4.6 (93\%) \\
10 & Red herring overload & 70.0\% & GPT-4o Mini (40\%) & Claude Haiku 4.5 (93\%) \\
11 & Framing/anchoring trap & 70.8\% & GPT-4o Mini (47\%) & GPT-4o (80\%) \\
12 & Self-defeating action & 72.5\% & GPT-4o Mini (47\%) & Multiple (80\%) \\
13 & Circular dependency & 73.3\% & GPT-4o (33\%) & Claude Sonnet (100\%) \\
14 & Naive physics error & 73.3\% & GPT-4o (40\%) & Claude Opus 4.6 (93\%) \\
15 & Embedded false premise & 75.8\% & GPT-4o Mini (7\%) & Haiku/Opus (100\%) \\
16 & Goal--means mismatch & 78.3\% & GPT-4o (27\%) & Haiku 4.5 (100\%) \\
17 & Temporal impossibility & 78.3\% & GPT-4o (13\%) & Haiku 4.5 (100\%) \\
18 & State/identity tracking & 80.0\% & GPT-4o Mini (60\%) & Haiku/Opus (93\%) \\
19 & Quantity/counting illusion & 81.7\% & GPT-4o Mini (47\%) & Claude Opus 4.6 (100\%) \\
20 & Scale/growth intuition fail. & 95.0\% & GPT-4o Mini (80\%) & Multiple (100\%) \\
\bottomrule
\end{tabular}
\end{table}

\section{Thinking Mode: Per-Category Deltas}
\label{app:thinking}

Table~\ref{tab:thinking_deltas} shows the per-category accuracy change when enabling extended thinking mode for both model pairs.

\begin{table}[ht]
\centering
\caption{Accuracy change (in \pp{}) from enabling extended thinking mode. Bold indicates changes $\geq$10\pp{} in magnitude.}
\label{tab:thinking_deltas}
\small
\begin{tabular}{@{}l r r@{}}
\toprule
\textbf{Category} & \textbf{GPT-5.4 $\Delta$} & \textbf{Claude Opus $\Delta$} \\
\midrule
Pragmatic/social intent & \textbf{$+$26.7} & \textbf{$+$33.3} \\
Answer hiding in plain sight & \textbf{$+$20.0} & \textbf{$+$20.0} \\
Wrong test conditions & \textbf{$+$20.0} & 0.0 \\
Implicit physical constraint & \textbf{$+$13.3} & $-$6.7 \\
Semantic scope trick & \textbf{$+$13.3} & \textbf{$-$20.0} \\
Default assumption hijack & 0.0 & \textbf{$+$20.0} \\
Broken/dead device self-ref. & $+$6.7 & \textbf{$+$20.0} \\
Wrong vantage point & $-$6.7 & $+$6.7 \\
Naive physics error & \textbf{$-$26.7} & 0.0 \\
Negation/exception logic & \textbf{$-$13.3} & 0.0 \\
Circular dependency & $-$6.7 & $-$6.7 \\
State/identity tracking & 0.0 & $-$6.7 \\
\bottomrule
\end{tabular}
\end{table}

\end{document}